\documentclass[letterpaper]{article} % DO NOT CHANGE THIS
\usepackage[submission]{aaai2026}  % DO NOT CHANGE THIS
\usepackage{times}  % DO NOT CHANGE THIS
\usepackage{helvet}  % DO NOT CHANGE THIS
\usepackage{courier}  % DO NOT CHANGE THIS
\usepackage[hyphens]{url}  % DO NOT CHANGE THIS
\usepackage{graphicx} % DO NOT CHANGE THIS
\usepackage{natbib}  % DO NOT CHANGE THIS AND DO NOT ADD ANY OPTIONS TO IT
\usepackage{caption} % DO NOT CHANGE THIS AND DO NOT ADD ANY OPTIONS TO IT
\usepackage{algorithm}
\usepackage{algorithmic}
\usepackage{amsmath}
\usepackage{newfloat}
\usepackage{listings}
\DeclareCaptionStyle{ruled}{labelfont=normalfont,labelsep=colon,strut=off} % DO NOT CHANGE THIS
\floatstyle{ruled}
\newfloat{listing}{tb}{lst}{}
\floatname{listing}{Listing}
\title{InstructUDrag: Joint Text Instructions and Object Dragging for Interactive Image Editing}
\author{
    Haoran Yu\textsuperscript{\rm 1},
    Yi Shi\textsuperscript{\rm 1}\thanks{Corresponding author}
}
\affiliations{
    \textsuperscript{\rm 1}School of Automation Science and Engineering,\\
    Faculty of Electronic and Information Engineering,\\
    Xi’an Jiaotong University, China\\
    HaoranYu@stu.xjtu.edu.cn, shiyi@mail.xjtu.edu.cn
}

\usepackage{bibentry}
\begin{document}

\maketitle

\begin{abstract}

Text-to-image diffusion models have shown great potential for image editing, with techniques such as text-based and object-dragging methods emerging as key approaches.   However, each of these methods has inherent limitations: text-based methods struggle with precise object positioning, while object dragging methods are confined to static relocation.   To address these issues, we propose InstructUDrag, a diffusion-based framework that combines text instructions with object dragging, enabling simultaneous object dragging and text-based image editing.   Our framework treats object dragging as an image reconstruction process, divided into two synergistic branches.  The moving-reconstruction branch utilizes energy-based gradient guidance to move objects accurately, refining cross-attention maps to enhance relocation precision.   The text-driven editing branch shares gradient signals with the reconstruction branch, ensuring consistent transformations and allowing fine-grained control over object attributes.   We also employ DDPM inversion and inject prior information into noise maps to preserve the structure of moved objects.   Extensive experiments demonstrate that InstructUDrag facilitates flexible, high-fidelity image editing, offering both precision in object relocation and semantic control over image content.
\end{abstract}


\section{Introduction}

Diffusion models have emerged as a dominant class of generative models due to their strong capability to produce diverse and high-quality samples \cite{ho2020denoising, song2020denoising, song2020score}, enabling a wide range of compelling applications \cite{meng2021sdedit, dong2023prompt, kumari2023multi}. Among these, image editing has emerged as a key application domain, spurring extensive research on leveraging diffusion models for controllable editing tasks \cite{hertz2022prompt, tumanyan2023plug, parmar2023zero, cao2023masactrl, kumari2023multi}. A popular approach for this task is text-conditioned editing, in which users guide image manipulation with natural language prompts.  While effective for global and local modifications, existing text-based methods often struggle with the precise spatial control of objects within the image, limiting their practical applicability.

\begin{figure}[ht!]
\centering 
\includegraphics[width=1\columnwidth]{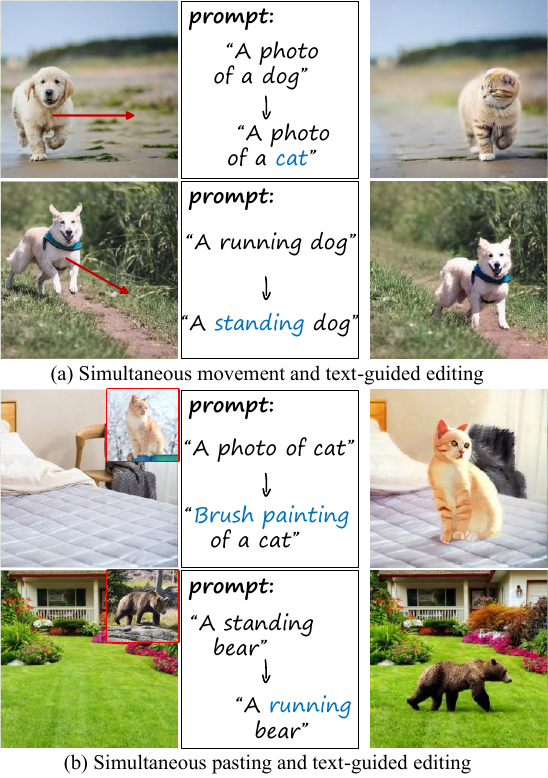} % Reduce the figure size so that it is slightly narrower than the column. Don't use precise values for figure width.This setup will avoid overfull boxes.
\caption{Our method allows users to specify a moving direction (or pasting position) of an object along with corresponding prompt words, Enabling text-driven editing of images while moving (or pasting).}
\label{fig1}
\end{figure}

Recently, drag-based diffusion editing methods, such as DragDiffusion \cite{shi2024dragdiffusion} and DragonDiffusion \cite{mou2023dragondiffusion}, have been proposed to address this limitation by enabling objects to be interactively dragged to target positions using pre-trained diffusion models. However, these methods are restricted to the relocation of objects and lack the flexibility to modify other image attributes simultaneously, such as style or action.   Moreover, preserving object structure and consistency during movement remains a significant challenge. In real-world scenarios, users often need to move objects while simultaneously modifying image details such as style, object appearance, or actions.

To overcome these challenges, we propose InstructUDrag, a novel framework that combines the strengths of both text-driven image editing and object dragging.  As illustrated in Figure~\ref{fig1}(a), InstructUDrag allows users to simultaneously drag objects to target positions while making detailed, text-guided edits to the image.  By treating object dragging as an image reconstruction process, we introduce a dual-branch architecture: the moving-reconstruction branch ensures structural consistency during object relocation, while the text-driven editing branch refines the image with prompt-driven modifications.  This synergy allows for greater accuracy and fidelity in both object movement and image transformation.  Crucially, the two branches interact via gradient guidance sharing.  In addition, to ensure accurate object relocation and resolve prompt ambiguities, we introduce non-target position mask learning, which optimizes the correspondence between the moved object and its textual description.  Furthermore, we leverage DDPM inversion and incorporate prior information into the noise maps to better preserve object features during movement, ensuring the generation of high-quality image output. Our method can also be extended to image pasting, as shown in Figure~\ref{fig1}(b).

Our contributions can be summarized as follows:  
\begin{itemize}
    \item Our method enhances the preservation of structural and textural details after object dragging, effectively removing attention traces from the object's source location.
    \item We treat object dragging as image reconstruction and propose a dedicated method, gradient guidance sharing, to enable simultaneous dragging and text-driven editing.
    % \item Extensive experiments and user study demonstrate the effectiveness of our approach in enabling simultaneous object dragging and text-driven editing.
    \item Extensive experiments demonstrate the effectiveness of our approach in enabling simultaneous object dragging and text-driven editing.
\end{itemize}

\section{Related work}

\subsubsection{Diffusion Models.}
Diffusion models are a class of generative models that progressively add noise to data and then learn to reverse this process to generate realistic samples. The Denoising Diffusion Probabilistic Model (DDPM) \cite{ho2020denoising} first showed that diffusion models can match state-of-the-art GANs \cite{goodfellow2020generative} in unconditional image generation. Later works, such as DDIM \cite{song2020denoising} and SDE \cite{song2020score}, refined the theory and improved sampling efficiency. To better align images with text prompts, GLIDE \cite{nichol2021glide} combined diffusion models with guidance techniques. Large-scale text-to-image diffusion models \cite{ramesh2022hierarchical, saharia2022photorealistic} further boosted image quality and controllability. Among them, Stable Diffusion (SD) \cite{rombach2022high} is especially popular for its efficiency. Unlike earlier approaches, SD projects images into a lower-dimensional latent space before adding noise, greatly reducing memory and compute costs. Our work builds on SD and fully leverages its generative power for controllable image editing.

\subsubsection{Text-to-Image Editing.}
Diffusion-based image editing has recently achieved remarkable results. For example, P2P \cite{hertz2022prompt} utilizes cross-attention control to edit images from one prompt to another. InstructPix2Pix \cite{brooks2023instructpix2pix}, Plug-and-Play \cite{tumanyan2023plug}, and Zero-Shot Image-to-Image Translation \cite{parmar2023zero} enable users to provide instructions or target prompts to manipulate real images toward desired edits. MasaCtrl \cite{cao2023masactrl} introduces mutual self-attention to query content from source images, achieving consistent edits. NTI \cite{mokady2023null} formulates editing as an image reconstruction problem to preserve structure and enable high-fidelity edits of real images. Variants based on DDPM/DDIM inversion \cite{miyake2025negative, huberman2024edit, wallace2023edict} further strengthen the source image’s influence by refining inversion formulations. Our method adopts DDPM inversion to better preserve the structure of images and objects during editing.

\subsubsection{Object Manipulation Editing.}
Object manipulation editing covers tasks such as object dragging, keypoint dragging, and object insertion, aiming to modify the spatial position, shape, or presence of objects. DragGAN \cite{parmar2023zero} first introduced drag-based interactive editing for consistent object manipulation. Building on this idea, DragDiffusion \cite{shi2024dragdiffusion} applied dragging to diffusion models, achieving improved performance and establishing a benchmark for subsequent work. DragonDiffusion \cite{mou2024diffeditor} and DiffEditor \cite{mou2024diffeditor} extend this by supporting both keypoint and object dragging, reformulating image editing as a gradient-based optimization process guided by an energy function. AnyDoor \cite{chen2024anydoor} and PaintByExample \cite{yang2023paint} train dedicated encoders for object insertion, which can also be adapted for dragging via inpainting. DiffOOM \cite{duan2025diffusion} proposes a framework for moving occluded objects. Move\&Act \cite{jiang2025move} enables simultaneous editing of an object’s action and its generated position. Unlike prior work, our method allows action modification during object dragging while also supporting local or global edits, providing more flexible and precise control.

\begin{figure*}[t]
\centering 
\includegraphics[width=1\textwidth]{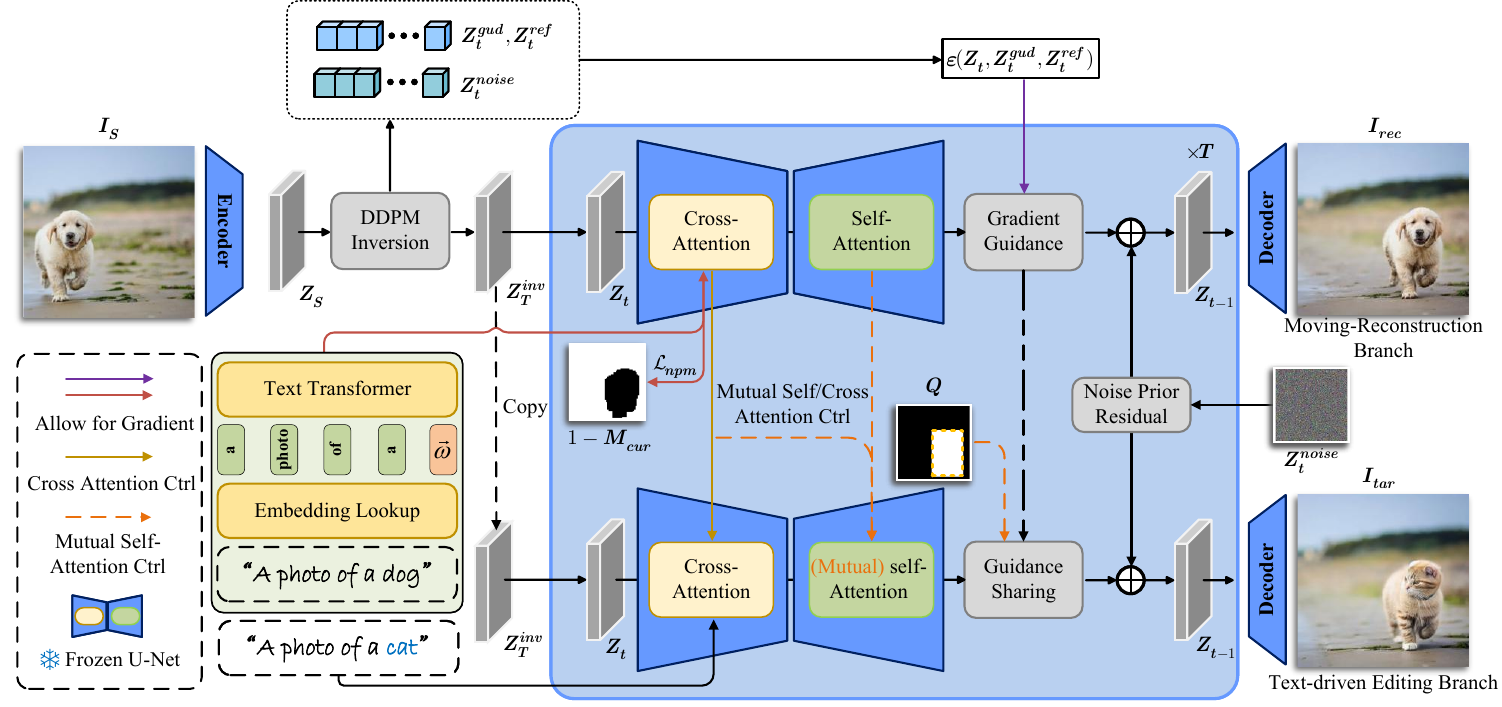} % Reduce the figure size so that it is slightly narrower than the column. Don't use precise values for figure width.This setup will avoid overfull boxes.
\caption{Pipeline of the proposed InstructUDrag. Our framework performs text-guided editing and object dragging simultaneously through two branches. The first branch, referred to as the moving-reconstruction branch, handles object relocation while preserving structural features. The second branch, called the text-driven editing branch,  performs text-guided image editing during object dragging, enabling synchronized semantic manipulation during dragging. Meanwhile, our method can flexibly adopt either cross-attention control or mutual self-attention control to enable more diverse text-driven image editing.}
\label{fig2}
\end{figure*}

\section{Methodology}

\subsection{Preliminaries}
\subsubsection{Score-based Gradient Guidance for Image Editing.}
The SDE (Stochastic Differential Equations) modeling for diffusion models was proposed in \cite{song2020score}, where estimating the score function in SDE is equivalent to noise prediction in diffusion models, the condition $y$ can be combined in a conditional score function, $i.e.$, $\nabla_{\mathbf{x}_t}\log q(\mathbf{x}_t|\mathbf{y})$. The conditional score function can be further derived as:\begin{equation} \begin{aligned} \nabla_{\mathbf{x}_t}\log q(\mathbf{x}_t|\mathbf{y})&=\log\left(\frac{q(\mathbf{x}_t|\mathbf{y})q(\mathbf{x}_t)}{q(\mathbf{y})}\right)\\&\propto\nabla_{\mathbf{x}_t}\log q(\mathbf{x}_t)+\nabla_{\mathbf{x}_t}\log q(\mathbf{y}|\mathbf{x}_t), \end{aligned} \label{eq:1} \end{equation} the first term represents the unconditional denoiser, the second term denotes the conditional gradient generated by an energy function, $i.e.$, $\mathcal{E}(\mathbf{x}_t,\mathbf{y})=\log{q(\mathbf{y}|\mathbf{x}_t)}$ to represent the distance between the text $y$ and $x_t$.

In subsequent image editing tasks, a series of score-based gradient guidance methods have been developed \cite{meng2021sdedit, mou2023dragondiffusion}. Specifically, DragonDiff constructs an energy function based on image feature correspondences and uses a pre-trained SD to estimate the energy function, $i.e.$, $\mathcal{E}(\mathbf{x}_t,\mathbf{x}_t^{inv})=\log{q(\mathbf{y}|\mathbf{x}_t)}$, where y is the editing target, to enable drag-style editing tasks. In our method, the moving-reconstruction branch adopts the DragonDiff framework to achieve object movement or pasting.

\subsection{Framework Overview}
As shown in Figure~\ref{fig2}, our framework takes the original image $I_s$, the source prompt $P_s$, and the movement direction of the target object as inputs, aiming to accurately perform object dragging while ensuring the edited image aligns with the target prompt $P_t$. Existing works often adopt a two-branch structure for reconstruction and editing \cite{cao2023masactrl, mokady2023null, shi2024dragdiffusion}. Inspired by current methods, we treat object dragging as an image reconstruction process with two synergistic branches. The first branch reconstructs the image while dragging the object to the target position, providing structural and content foundations for subsequent edits. The second branch builds on this and performs prompt-guided fine-grained editing for precise semantic control. The first branch provides gradient guidance that is shared with the second branch for coherent editing. To further enhance relocation accuracy and reduce prompt ambiguity, we introduce non-target position mask learning to refine the prompt representation of the moved object. Finally, we leverage DDPM inversion to enable faithful reconstruction and flexible editing.

\begin{figure}[t]
\centering 
\includegraphics[width=1\columnwidth]{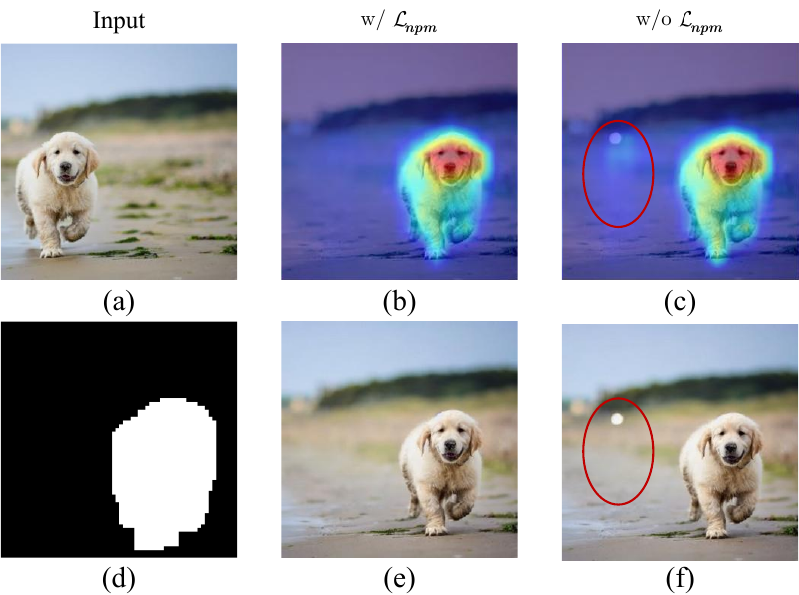} % Reduce the figure size so that it is slightly narrower than the column. Don't use precise values for figure width.This setup will avoid overfull boxes.
\caption{Effectiveness of the non-target position mask loss $\mathcal{L}_{npm}$. (a) is the input image, and (d) is the mask $\mathbf{M}_c$ of the target position. (b) and (c) are the attention maps corresponding to the object's prompt with/without $\mathcal{L}_{npm}$, (e) and (f) are the images output by the moving-reconstruction branch with/without $\mathcal{L}_{npm}$.}
\label{fig3}
\end{figure}

\subsection{Non-target Position Mask Learning}
During the object dragging editing process, we observe that cross-attention traces often persist at the object’s original position, as illustrated in the cross-attention map for the word “dog” in Figure~\ref{fig3}(c), where the red circle highlights the residual attention at its initial location. This phenomenon tends to degrade the quality of the generated images, as illustrated in Figure~\ref{fig3}(f), and may further interfere with subsequent text-based editing tasks.

To address this issue, inspired by textual inversion and dynamic prompt learning \cite{gal2022image, yang2023dynamic}, we proposed non-target position mask learning (NPML). This method aims to refine the cross-attention map associated with the moved object by learning to suppress attention outside the target region defined by the prompt at each denoising timestep, thereby eliminating residual traces at the original location.

Specifically, during object dragging, we generate a mask $M_c$ (Figure~\ref{fig3}(d)), which represents the target region of the moved object. NPML minimizes the overlap between the cross-attention map for the corresponding prompt and the non-target regions outside $M_c$. To this end, we introduce the following loss function $\mathcal{L}_{npm}$:\begin{equation} \begin{aligned} \mathcal{L}_{npm}=\lambda_{c}\cdot\cos\left(\mathbf{A}_{t}^{w}, 1-\mathbf{M}_{c}\right) + \lambda_{i}\cdot\frac{\mathbf{A}_{t}^{w}\cdot(1-\mathbf{M}_{c})}{\sum_{j=1}^{D}\mathbf{A}_{t,j}}, \end{aligned}\end{equation} where $\cos(\cdot)$ is the calculation of cosine similarity, $w$ is the learnable word embedding for the moving object, and $\mathbf{A}_{t}^{w}$ is its cross-attention map at timestep $t$. $\lambda_c$ and $\lambda_i$ are two hyper parameters. The first term encourages low similarity between the attention map and the non-target region, while the second term computes the intersection-over-union (IoU) to explicitly penalize residual attention outside the mask. By minimizing $\mathcal{L}_{\text{npm}}$, the word embedding $w$ is optimized to suppress residual attention outside the target region, thereby refining the cross-attention map for the moved object, as demonstrated in Figure~\ref{fig3}(b).

\subsection{Editing with Gradient Guidance Sharing}

\begin{figure}[t]
\centering 
\includegraphics[width=1\columnwidth]{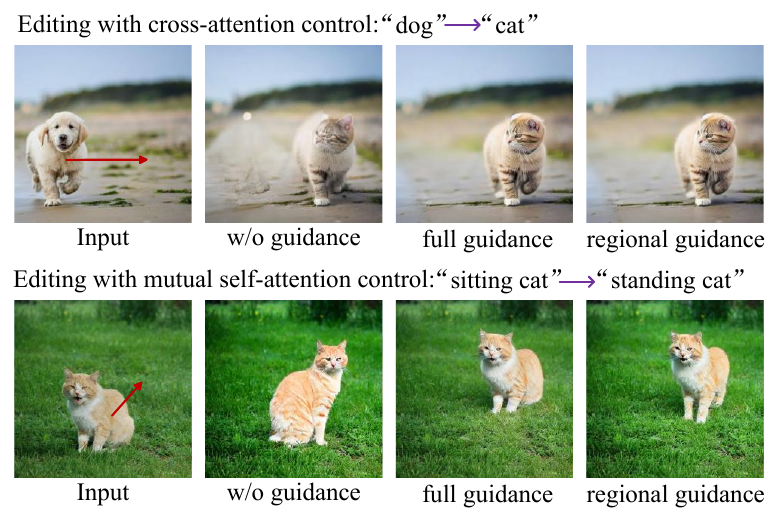} % Reduce the figure size so that it is slightly narrower than the column. Don't use precise values for figure width.This setup will avoid overfull boxes.
\caption{First row: text-driven editing branch using cross-attention control, with/without gradient guidance sharing. Second row: text-driven editing branch using mutual self-attention control, with/without gradient guidance sharing.}
\label{fig4}
\end{figure}

\begin{figure*}[t]
\centering 
\includegraphics[width=1\textwidth]{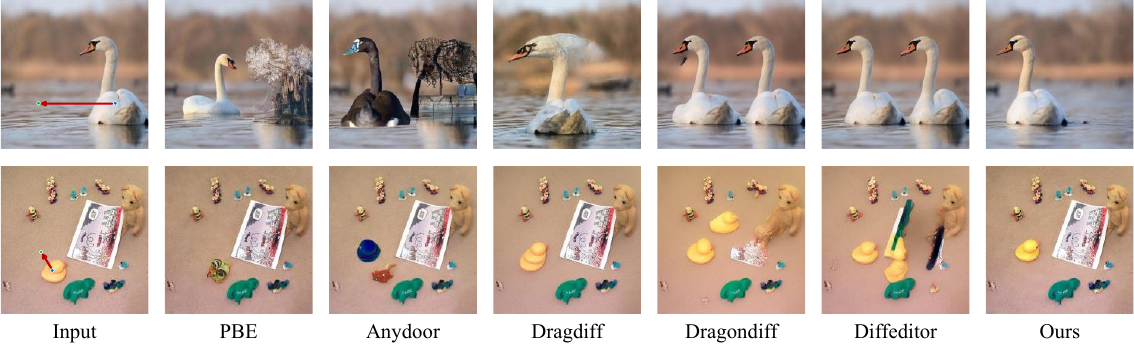} % Reduce the figure size so that it is slightly narrower than the column. Don't use precise values for figure width.This setup will avoid overfull boxes.
\caption{Qualitative comparison on object dragging. The red arrow indicates the direction of the object dragging. The source and target locations are denoted by blue and green points, respectively.}
\label{fig5}
\end{figure*}

Dragondiff \cite{mou2023dragondiffusion} constructs an energy function, transforming image editing into gradient guidance in diffusion sampling. The energy function consists of two parts: the editing term $\mathcal{E}_{edit}$ and the content consistency term $\mathcal{E}_{content}$.  To enable more fine-grained control over different regions, we decouple contrast loss $\mathcal{L}_{\text{contrast}}$ and inpainting loss $\mathcal{L}_{\text{inpaint}}$ from $\mathcal{E}_{\text{content}}$, forming a new energy function $\mathcal{E}_{\text{inpaint}}$ and apply regional energy functions to three distinct areas. Thus, the conditional term in Eq.~\ref{eq:1} can be formulated as:\begin{equation} \begin{aligned} \nabla_{\mathbf{z}_{t}}\log q(\mathbf{y}|\mathbf{z}_{t})=&\mathbf{M}_{c}\cdot\nabla_{\mathbf{x}_{t}}\mathcal{E}_{edit}+\mathbf{M}_{v}\cdot\nabla_{\mathbf{x}_{t}}\mathcal{E}_{inpaint}\\&+(1-(\mathbf{M}_{c} \cup \mathbf{M}_{v}))\cdot\nabla_{\mathbf{x}_{t}}\mathcal{E}_{content}, \end{aligned}\end{equation} $\mathbf{M}_{c}$ and $\mathbf{M}_{v}$ are the masks for the target position and the source position, respectively.

In diffusion based training-free image editing, there are typically two branches: reconstruction and editing. Dragdiff \cite{shi2024dragdiffusion} trains a LoRA for precise image reconstruction, while NTI \cite{mokady2023null} optimizes unconditional embeddings to achieve approximately precise reconstruction results. Inspired by this two-branch editing paradigm, our method treats object dragging as a reconstruction process. As shown in the model structure diagram of Figure~\ref{fig2}, our framework consists of a moving-reconstruction branch and a text-driven editing branch.  

In the moving-reconstruction branch, gradient guidance is employed for object relocation and image reconstruction. The gradient guidance at each timestep is stored and shared with the text-driven editing branch. By leveraging this shared guidance, the text-driven branch enables simultaneous object dragging and text-based editing. Text-based editing is performed using cross-attention control \cite{hertz2022prompt} or mutual self-attention control \cite{cao2023masactrl}.

\subsection{Inversion with DDPM Noise Prior}

In conventional DDIM inversion \cite{song2020denoising}, a deterministic mapping from a single noise map to the generated image is used, which often fails to faithfully preserve fine details of the input image. While DDPM inversion \cite{huberman2024edit} extracts T + 1 noise vectors from the generation process, "imprints" the image more strongly onto the noise maps.

To preserve the structure of moving objects during object dragging, we propose to add prior information to extracted noise maps $ \{z_1, ... ,z_T\} $ of DDPM inversion:\begin{equation} \begin{aligned} z_t^{edit} = (1-\mathbf{M}_c) \cdot z_t + \mathbf{M}_c \cdot shift(z_t, dx, dy) \end{aligned}\end{equation} $shift(z_t, dx, dy)$ denotes the feature map obtained by shifting the noise feature map $z_t$ to the target position. Our method can also be extended to object pasting, where the feature map to be shifted is the pasting object. By introducing noise prior during DDPM inversion, our method more effectively preserves both structural and textural details of the edited images.

To further refine the editing process, we integrate this noise prior with region-specific gradient guidance sharing. Specifically, when mutual self-attention control is employed, we generate a mask $Q$ that defines a rectangular region centered on the moving target. Gradient guidance sharing is then applied only outside this region, as applying it within $Q$ would overly preserve the structure of the target object, hindering its intended modification, as demonstrated in the second row of Figure~\ref{fig4}. In contrast, when cross-attention control is used, we adopt full gradient guidance sharing to maintain the global structure of the image. The effects of these different gradient guidance strategies are illustrated in Figure~\ref{fig4}.

\section{Experiments}

\begin{figure*}[t]
\centering 
\includegraphics[width=1\textwidth]{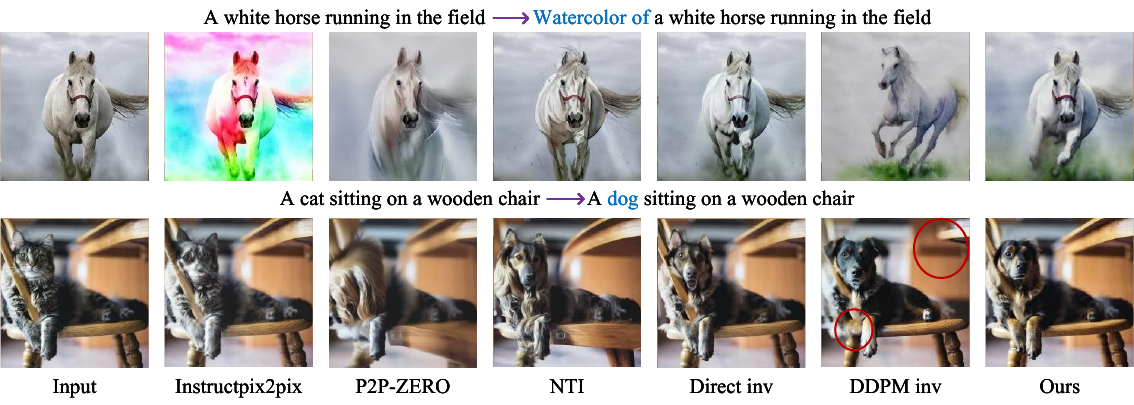} % Reduce the figure size so that it is slightly narrower than the column. Don't use precise values for figure width.This setup will avoid overfull boxes.
\caption{Visual comparison of different methods for modifying only through text-driven editing.}
\label{fig6}
\end{figure*}

% \begin{table*}[h]
%     \centering
    
%     \begin{tabular}{l|c|c|c}
%         \hline\noalign{\hrule height 0.2 pt}\noalign{\vskip 2 pt}
%         Method       & Foreground $\uparrow$ & Traces $\downarrow$ & Realism $\downarrow$ \\
%         \noalign{\vskip 2 pt}\hline\noalign{\vskip 2 pt}
%         PBE[CVPR2023]          & 0.824   & 0.674   & 0.0266              \\
%         Anydoor[CVPR2024]      & 0.852   & 0.673   & 0.0271              \\
%         DragDiff[CVPR2024]     & 0.802   & 0.676   & 0.0235              \\
%         DragonDiff[ICLR2024]   & 0.892   & 0.711   & 0.0239              \\
%         DiffEditor[CVPR2024]   & 0.902   & 0.716   & 0.0258              \\
%         Ours         & \textbf{0.910}   & \textbf{0.633}   & \textbf{0.0232}              \\
%         \noalign{\vskip 2 pt}\hline\noalign{\hrule height 0.2 pt}
    
%     \end{tabular}
%     \caption{Quantitative comparison on object dragging. Our method outperforms the compared methods in foreground similarity, object traces, and realism.}
%     \label{table1}
% \end{table*}

\subsection{Implementation Details}

We use Stable Diffusion \cite{rombach2022high} with checkpoint v1.5 for image editing. The hyperparameter values for $\lambda_c$ and $\lambda_i$ in Eq. (3) are both set to 0.5. The DDPM inversion inference steps we adopt are 100 steps, and the parameter $T_{skip}$, which controls the adherence to the input image, is set to 15. Meanwhile, the classifier-free guidance scales for the moving-reconstruction branch and the prompt-editing branch are 3.5 and 7.5, respectively. All experiments were conducted on an NVIDIA 3090 GPU, and all test images were processed to 512×512 resolution.

% \bigskip
% \noindent Thank you for reading these instructions carefully. We look forward to receiving your electronic files!

\begin{table}[t]
    \centering
    
    \begin{tabular}{l|c|c|c}
        \hline\noalign{\hrule height 0.2 pt}\noalign{\vskip 2 pt}
        Method       & Foreground $\uparrow$ & Traces $\downarrow$ & Realism $\downarrow$ \\
        \noalign{\vskip 2 pt}\hline\noalign{\vskip 2 pt}
        PBE          & 0.824   & 0.674   & 0.0266              \\
        Anydoor      & 0.852   & 0.673   & 0.0271              \\
        DragDiff     & 0.802   & 0.676   & 0.0235              \\
        DragonDiff   & 0.892   & 0.711   & 0.0239              \\
        DiffEditor   & 0.902   & 0.716   & 0.0258              \\
        Ours         & \textbf{0.910}   & \textbf{0.633}   & \textbf{0.0232}              \\
        \noalign{\vskip 2 pt}\hline\noalign{\hrule height 0.2 pt}
    
    \end{tabular}
    \caption{Quantitative comparison on object dragging. Our method outperforms the compared methods in foreground similarity, object traces, and realism.}
    \label{table1}
\end{table}

\subsection{Comparison on Object Dragging}

To evaluate the performance of our method in object dragging, we compare the dragging-only images generated by the moving-reconstruction branch in our method with those generated by other most relevant available methods. Paint-By-Example (PBE) \cite{yang2023paint} and Anydoor \cite{chen2024anydoor} are methods for object pasting, we achieve the object dragging by applying inpainting technology at the original position of the target object and pasting it at the target position. DragDiffusion \cite{shi2024dragdiffusion} is a point-based dragging method that enables object movement by selecting multiple dragging points on the target object. DragonDiffusion \cite{mou2023dragondiffusion} and DiffEditor \cite{mou2024diffeditor} directly apply to object dragging.

% We annotated 125 images from the web. For each image, we used SAM \cite{kirillov2023segment} to obtain the visible mask $\mathbf{M}_{v}$ of the moving object at its original position, and each image was annotated with "a photo of <object>". We calculated 8 different target dragging positions, resulting in a total of 1000 testing cases. We conducted quantitative comparisons using 3 different metrics: foreground similarity, object traces, and realism. Foreground similarity indicates whether the object is accurately dragged to the target position. In the edited image, we cropped a tight box around $\mathbf{M}_{c}$ to get $\mathbf{I}_{c}$, and in the original image, a tight box around $\mathbf{M}_{v}$ to get $\mathbf{I}_{s}$. We extracted the features of these two crops using DINOv2 \cite{oquab2023dinov2} to calculate the cosine similarity. A higher similarity score indicates that the object is successfully dragged to the target position with better feature preservation.Object traces refer to whether the object is erased from its original position without leaving any traces. We cropped a tight box around $\mathbf{M}_{v}$ in the edited image to obtain $\mathbf{I}_{e}$, and also used DINOv2 to extract features from $\mathbf{I}_{e}$ and $\mathbf{I}_{s}$ for cosine similarity calculation. A lower similarity score indicates a better erasure effect of the object at the original position.Additionally, we calculated the KID \cite{binkowski2018demystifying} score between 500 real images and the edited images to measure the realism of the edited images. 

We annotated 125 web-sourced images and used SAM \cite{kirillov2023segment} to generate the visible mask $\mathbf{M}_{v}$ of the object at its original position. We then defined 8 target positions per image, resulting in a total of 1,000 test cases.For quantitative evaluation, we used three metrics:  foreground similarity, object traces, and realism. Foreground similarity evaluates the accuracy of object relocation to the target location. Specifically, we crop a tight box around the target mask $\mathbf{M}_{c}$ in the edited image to obtain $\mathbf{I}_{c}$, and a tight box around the source mask $\mathbf{M}_{v}$ in the original image to obtain $\mathbf{I}_{s}$. We then extract features using DINOv2 \cite{oquab2023dinov2} and compute cosine similarity, a higher score indicates better feature preservation after dragging.Object traces measure how well the object is removed from its original position. We crop a tight box around $\mathbf{M}_{v}$ in the edited image to obtain $\mathbf{I}_{e}$, extract features with DINOv2, and compute the cosine similarity between $\mathbf{I}_{e}$ and $\mathbf{I}_{s}$. A lower score indicates fewer residual traces.Finally, we compute the KID score \cite{binkowski2018demystifying} between 500 real images and the edited results to assess the realism of the generations.

\begin{figure*}[t]
\centering 
\includegraphics[width=1\textwidth]{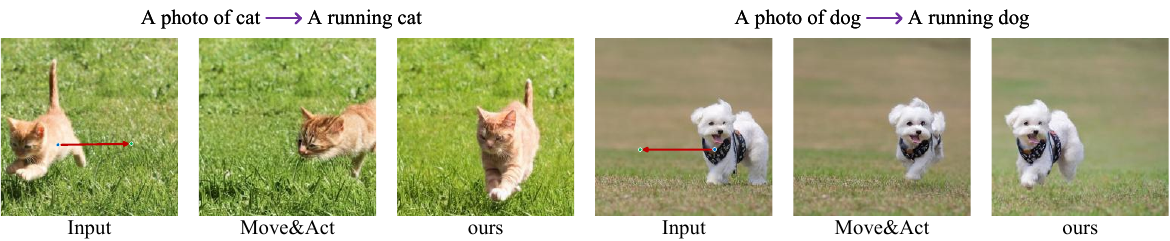} % Reduce the figure size so that it is slightly narrower than the column. Don't use precise values for figure width.This setup will avoid overfull boxes.
\caption{Visual comparison between our method and Move\&Act.}
\label{fig7}
\end{figure*}

\begin{figure*}[t]
\centering 
\includegraphics[width=1\textwidth]{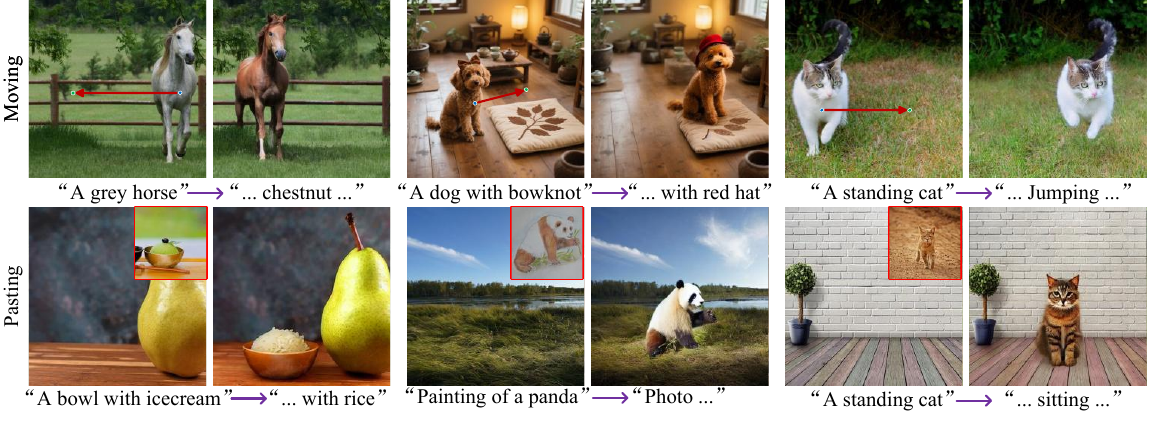} % Reduce the figure size so that it is slightly narrower than the column. Don't use precise values for figure width.This setup will avoid overfull boxes.
\caption{Visualization of our method on multiple editing approaches.}
\label{fig8}
\end{figure*}

As shown in Table~\ref{table1}, PBE and Anydoor achieved low foreground similarity scores due to issues in preserving the appearance and features of the moved object, and the overall realism of the images edited by these two methods is also low. DragDiffusion performed poorly in terms of foreground similarity because it is mainly used for point dragging rather than the overall movement of objects. DragonDiffusion and DiffEditor struggled with object traces, as they failed to effectively erase the object from its original position. Our method outperformed the baselines in all three metrics, which can also be seen from the qualitative comparison in Figure~\ref{fig5}.

\subsection{Comparison on Text-based Editing}

To demonstrate that our designed gradient guidance sharing has the ability to preserve essential content and supports prompt-based image editing, we set the movement distance of the object in the Moving-Reconstruction Branch to 0, turning this branch into a pure reconstruction branch, so as to test the image editing performance of our method.
As shown in Figure~\ref{fig6}, we compare our method with existing text editing methods, including the trained method Instructpix2pix \cite{brooks2023instructpix2pix}, the gradient-based text method P2P-Zero \cite{parmar2023zero} and the background-preserving methods NTI \cite{mokady2023null} and Direct inversion \cite{ju2023direct}. Since our method adopts DDPM inversion \cite{huberman2024edit} as the inversion method, we also compared our method with the one using only DDPM inversion. Except for Instructpix2pix and P2P-Zero, all methods employ cross-attention control for prompt editing. When DDPM inversion is used without gradient guidance sharing, the editing results often have large deviations compared with the original image. Meanwhile, our method produces competitive editing results compared with NTI and Direct inversion, which further verifies that our method supports prompt-based image editing. Our method can also be regarded as a background-preserving method, and the object dragging process can be considered as a form of background preservation. Therefore, our method can perform both object dragging and prompt editing simultaneously, which constitutes a significant advantage over other editing methods.

\begin{table}[t]
    \centering
    
    \begin{tabular}{l|c|c}
        \hline\noalign{\hrule height 0.2 pt}\noalign{\vskip 2 pt}
        Method       & CLIP-Score $\uparrow$ & Realism $\downarrow$ \\
        \noalign{\vskip 2 pt}\hline\noalign{\vskip 2 pt}
        Move\&Act          & 0.3098   & \textbf{0.0545}          \\
        Ours         & \textbf{0.3141}   & 0.0572        \\
        \noalign{\vskip 2 pt}\hline\noalign{\hrule height 0.2 pt}
    
    \end{tabular}
    \caption{Quantitative comparison on overall performance.}
    \label{table2}
\end{table}

\subsection{Overall Performance}

We compared the overall performance of our method with Move\&Act \cite{jiang2025move}. Since Move\&Act only supports consistent image editing, we used mutual self-attention control to modify the actions of animals in the evaluation for a fair comparison.We tested 200 image pairs from the Move\&Act dataset by simultaneously changing the positions and postures of animals.For quantitative evaluation, we adopted two metrics: CLIP Score \cite{radford2021learning} and KID. The CLIP Score measures text–image alignment, where a higher value indicates better consistency with the input prompt. The KID assesses the visual realism of the generated images. The quantitative results are presented in Table~\ref{table2}. Our method achieved a higher CLIP Score, demonstrating better alignment with user instructions, but performed slightly worse than Move\&Act in terms of realism. Qualitative examples comparing both methods are shown in Figure~\ref{fig7}. In contrast, Move\&Act may suffer from issues such as incomplete feature preservation after movement or failure to accurately place objects at the desired locations.

Another major advantage of our method over Move\&Act is that our method can not only perform consistent editing while object dragging, but also support diverse text-based editing, such as style modification and local detail refinement. Furthermore, our method can be extended from object dragging to object pasting, enabling more flexible object insertion, including the seamless blending of stylistically diverse pasted objects with the surrounding scene. Additional qualitative results are presented in Figure~\ref{fig8}.

\begin{table}[t]
    \centering
    
    \begin{tabular}{l|c|c|c}
        \hline\noalign{\hrule height 0.2 pt}\noalign{\vskip 2 pt}
        Method       & Traces $\downarrow$ & Foreground $\uparrow$ & CLIP-Score $\uparrow$ \\
        \noalign{\vskip 2 pt}\hline\noalign{\vskip 2 pt}
        Ours         & \textbf{0.713} & \textbf{0.920} & \textbf{0.3148}              \\
        \noalign{\vskip 2 pt}\hline\noalign{\vskip 2 pt}
        w/o GGS      & 0.878 & 0.887 & 0.3091              \\
        w/o NPML     & 0.717 & 0.919 & 0.3141              \\
        w/o DNP   & 0.714 & 0.886 & 0.3144              \\
        w/o DREF   & 0.781 & 0.914 & 0.3139              \\
        \noalign{\vskip 2 pt}\hline\noalign{\hrule height 0.2 pt}
    
    \end{tabular}
    \caption{Ablation Study. We ablate the following components of our method: (1) w/o Gradient Guidance Sharing (GGS), (2) w/o Non-target Position Mask learning (NPML), (3) w/o DDPM Noise Prior (DNP), (4) w/o Decoupled Region Energy Function (DREF) }
    % \caption{Ablation Study.}
    \label{table3}
\end{table}

\subsection{Ablation Study}
We conducted an ablation study on the following components of our method. Similarly, 200 image pairs from the Move\&Act dataset were tested, with the positions and postures of the animals in the images changed simultaneously. As can be seen in Table~\ref{table3}: (1) Gradient Guidance Sharing (GGS): After removing GGS from our method, all three metrics dropped significantly, as it serves as a crucial bridge connecting the two branches in our method. (2) Non-target Position Mask learning (NPML): Not using NPML resulted in a higher traces score and a lower CLIP score, which is because NPML functions to eliminate the attention traces of the object at its original position and strengthen the concept of the object at the target position. (3) DDPM Noise Prior (DNP): Removing DNP is detrimental to foreground preservation, thus leading to a significantly lower foreground score. (4) Decoupled Region Energy Function (DREF): When DREF is replaced with the original energy function, the trace score increases. This is because our method introduces a decoupled energy term $\mathcal{E}_{\text{inpaint}}$, specifically designed for inpainting. By independently controlling the scale of $\mathcal{E}_{\text{inpaint}}$, our approach enables more effective removal of residual traces at the object's original location.

\section{Conclusion}
% We propose a training-free framework that enables flexible and precise control over both object dragging and text-driven editing. Our method treats object dragging as image reconstruction and shows that sharing gradient guidance enables simultaneous object dragging and text-driven editing. We also adopt a series of strategies to eliminate the traces of the object at its original position after it is moved and to enhance the concept of the target position. Our method can be combined with cross-attention control or mutual self-attention control to achieve more diverse image editing, such as modifying the target's action, local details, or image style while moving the object. Additionally, our method can be extended to object pasting.  Moreover, our method can be extended to object pasting.

We propose a training-free framework that enables flexible and precise control over both the position of objects and image semantics. Our method treats object dragging as an image reconstruction process, and shows that sharing gradient guidance enables simultaneous object dragging and text-based editing. We also adopt a series of strategies to eliminate the traces of the object at its original position and to enhance the concept of the target position. Moreover, our method can be naturally extended to support object pasting.  Future work will focus on extending to dynamic environments and incorporating real-time feedback mechanisms, further enhancing its utility for interactive applications.

% We propose a training-free framework that enables flexible and precise control over both object dragging and text-driven image editing. Our method treats object dragging as an image reconstruction process, and shows that sharing gradient guidance enables simultaneous object dragging and text-driven image editing. We also adopt a series of strategies to eliminate the traces of the object at its original position and to enhance the concept of the target position. Additionally, our method can be extended to object pasting. Future work will focus on extending InstructUDrag to dynamic environments and incorporating real-time feedback mechanisms, further enhancing its utility for interactive applications. 

% We propose a training-free framework that enables flexible and precise control over both object dragging and text-driven image editing. By formulating object dragging as an image reconstruction process, our method demonstrates that sharing gradient guidance between branches allows both operations to be performed simultaneously. We also adopt a series of strategies to eliminate the traces of the object at its original position after it is moved and to enhance the concept of the target position. Moreover, our framework can be naturally extended to support object pasting. Future work will focus on extending to dynamic environments and incorporating real-time feedback mechanisms, further enhancing its utility for interactive applications.

\bibliography{aaai2026}

\end{document}